\documentclass{article}
\usepackage{hyperref}
\usepackage{amsmath}
\usepackage{amsfonts}
\usepackage{amssymb}

\title{Possible Mechanisms for Neural Reconfigurability	and their
Implications\protect\footnotetext{
\noindent This paper is the long version of talk presented at Snowbwird Learning Workshop, April 2012\protect\cite{original}.
}
}
\author{Thomas M. Breuel \\ University of Kaiserslautern}
\date{}
\begin{document}

\maketitle
\begin{abstract}
The paper explores a biologically and evolutionarily plausible neural
architecture that allows a single group of neurons, or an entire cortical
pathway, to be dynamically reconfigured to perform multiple, potentially
very different computations. We observe that reconfigurability can
account for the observed stochastic and distributed coding behavior of
neurons and provides a parsimonious explanation for timing phenomena in
psychophysical experiments. It also shows that reconfigurable pathways
correspond to classes of statistical classifiers that include decision
lists, decision trees, and hierarchical Bayesian methods. Implications
for the interpretation of neurophysiological and psychophysical results
are discussed, and future experiments for testing the reconfigurability
hypothesis are explored.
\end{abstract}

\section{Introduction}

A common underlying assumption of much of modern neurophysiology
and functional brain imaging is that the brain is divided into
areas with different, identifiable functions, and that within
each area, neurons perform specific computations.

Consider a pathway like the visual pathway
(V1, V2, V4, IT).
Commonly, such a pathway is assume to compute a sequence
of representations of the sensory input at increasing levels
of abstraction.
However, in the visual pathway, despite extensive attempts
to identify the nature of representation in V4 and IT, not
clear representation has been identified.

The {\em reconfigurability hypothesis} proposed in \cite{original}
states that we should view such a pathway not as a neural
network with a fixed function, but instead as a reconfigurable
network that performs different computational tasks at
different times.
In effect, the pathway is temporally multiplexed between
different functions.

A reconfigurable pathway can be in one of many
different states ({\em configurations}).
In different configurations, areas may perform different
computations and generate different representations;
in particular, the activities of individual neurons
may be unrelated in different configurations.
In a sense, we think of neural pathways as being similar
to digital FPGAs (field programmable gate arrays).

Reconfiguration of pathways is postulated to be under
control of one or more brain areas outside the pathway.
Furthermore, we generally consider pathway configurations
in which the pathway itself operates largely like
a feed-forward network.  We will see that combining
feed-forward networks with reconfigurability results in
additional computational power that allows feed-forward
networks to perform some of the functions usually
thought of as requiring recurrent architectures.

As we will see, reconfigurable networks are easy to
create with simple, biologically plausible mechanisms.
We will argue that they likely confer a significant
evolutionary advantage on animals because neurons
require large amounts of energy to maintain and
reconfigurability permits an organism to perform more
diverse and complex computations with the same number
of neurons.
In addition, reconfigurability provides a simple explanation
for a number of puzzling observations in neurophysiology
and psychophysics that currently have no widely accepted,
specific computational models, such as stochastic neural
responses and timing behaviors.

A second important issue in understanding the function
of neural networks has been the question of how recurrency is involved
in neural computations.
Neural network models (spiking or McCulloch-Pitts-style) are
generally divided into two classes:
feed-forward networks and recurrent networks.
Feed-forward networks are widely used and well studied;
they compute non-linear functions that are commonly
used as decision functions in classifiers, or
function approximators.
Recurrent networks have temporal state and dynamics.
There is a wide variety of models, such as the
Hopfield model\cite{HopfieldModel}, Bayesian message passing
algorithms, liquid or echo state networks\cite{EchoState},
and Boltzmann machines\cite{Boltzmann1,Boltzmann2}.
In addition, special-purpose recurrent networks, such
as networks for attentional selection\cite{AttentionalSelection} and
segmentation have been proposed.

Numerous effects have been observed, both neurophsiologically
and psychophysically, that cannot be explained with
simple feedforward models, yet explanation in terms
of biologically plausible recurrent networks has been difficult.
We will see that reconfigurability not only results in
more efficient utilization of neural hardware, but
it also extends the computational capabilities of
feed-forward neural networks significantly.

In the rest of the paper, we will first formulate
and formalize a simple model of reconfigurable
feed-forward networks, then examine how such models
related to neurophysiology, psychophysics, and
statistical learning.

\section{Reconfigurable Feed-Forward Models}

Although reconfigurable models can be formulated
for many different kinds of networks (including
spiking neural networks and recurrent networks),
for concreteness, let us consider a simple model
primarily based on feed-forward models for most
of the rest of the paper.

\subsection{Multiplexed Feed-Forward Models}

Consider a feed-forward model with linear combination
of input values, followed by a non-linearity.
Let the activity of unit $j$ at layer $k$ be represented as $x^{(k)}_j$,
let the weight matrix at layer be $M^{(k)}_{i,j}$,
and let the nonlinearity be given by $\sigma(\cdot)$
(we assume that this is an element-wise non-linearity,
like the sigmoid function, and/or a normalization function
like $f(x)=x/||x||$).  Then a feed-forward model is given by:

\begin{equation}
x^{(k+1)} = \sigma\left( \sum M^{(k)}_{i,j} x^{(k)}_j \right)
\end{equation}

\noindent Reconfigurability postulates an additional set of parameters
$\lambda_{r}$ and replaces the feed-forward equations with:

\begin{equation}\label{eq-multiplexed}
x^{(k+1)} = \sigma\left( \sum_r \lambda_r \sum_j M^{(k,r)}_{i,j} x^{(k)}_j \right)
\end{equation}

\noindent (Additional nonlinearities could be introduced inside the outer sum.)

We are primarily considering models in which the pathways
are in one configuration $r$ of $R$ discrete configurations.
That is, $\lambda_r\in\{0,1\}^R$, $\sum_r \lambda_r = 1$
(more general models of the form $\lambda_r\in\mathbb{R}$ are also possible).

From these equations, it is easy to see that for each $r\in\{1,...,R\}$,
the network computes completely the result of applying a 
different feed-forward network to the inputs.

Let us call these kinds of feed-forward models a {\em multiplexed feed-forward
model}, since the same neural pathway can be used to implement multiple
different (and potentially unrelated) computations in sequence.
This kind of mechanism allows the same number of neurons to perform
many more computations compared to a non-multiplexed pathway.
It is attractive because it is easily implemented via a biologically
plausible mechanism.

\subsection{Sequential Control of Configurations}

Multiplexed feed-forward models are one component of reconfigurable
network models.  The second component we need is a control mechanism that
chooses configurations.  In the most general case, a control mechanism
might consist of a recurrent network receiving as inputs sensory inputs,
internal state, and outputs from the multiplexed pathway.
But, perhaps surprisingly, we can perform many kinds of useful computations
with much more limited control mechanisms.  These kinds of more limited
control mechanisms will be useful in formulating hypotheses about the
timing behavior in psychophysical experiments,
and they also provide a plausible evolutionary path by which more complex
control mechanisms may have evolved.
To describe these control mechanisms, consider a {\em task} to be solved,
like a sequence of presentations and responses in a psychophysical vision
experiment.

\begin{itemize}

\item[Type I] The simplest control mechanism is to choose a {\em fixed configuration}
at the beginning of a task and then operate with that configuration
throughout the duration of the task.
This is the kind of setting in which we can explain results such as
fast feed-forward object recognition\cite{PoggioObject}.
The configuration itself is chosen by mechanisms outside the reconfigurable
pathway and it is chosen independently of its outputs.

\item[Type II] Configurations may also be chosen using {\em hierarchical control}.
In that case, a separate pathway performs analysis of sensory input
(say, itself using a feed-forward architecture), and its output then
controls the configuration of the reconfigurable pathway.

\item[Type III] Configurations may be {\em sequentially multiplexed} during
a task. That is, the parameters $\lambda_r$ cycle through
a fixed set of values, corresponding to different configurations
of the pathway.
The output of the reconfigurable pathway switches between
the results of different computations.
These may then be integrated by a subsequent feed-forward
network (possibly after some temporal integration).

\item[Type IV] Configurations may be {\em sequentially tried out}
during a task, where each configuration attempts to complete the task;
if a configuration is successful, the task is completed and the
sequence of configurations starts again at the beginning.

\item[Type V] Configurations {\em depend on the output} of the
reconfigurable pathway, but the only information that is recurrent
is the set of parameters $\lambda_r$.
The overall output from the pathway is computed possibly by
an additional feed-forward computation operating on the output of the
configurable pathway, as in Type III.

\item[Type VI] Configurations depend on the output of the reconfigurable
pathway, but in addition, other information is propagated recurrently.
For example, the control mechanism may cause activations to be
buffered or stored inside the pathway.

\end{itemize}

Except for Type~I control, all these mechanisms assume that the
successful completion of some recognition requires the use of
multiple configurations, and that the integration of the results
from these configurations yields the final answer to the perceptual
problem that is being solved during the task.
As a concrete example, consider a task in which subjects need
to classify either outdoors scenes or recognize letters;
solving this task would run through a sequence of configurations
specialized for recognizing outdoors scenes and a sequence of
configurations specialized for recognizing letters.
Depending on which control mechanism is used, we would expect
different kinds of timing behaviors.
Each configuration would indicate whether it successfully
interpreted the input and what class the input belongs to.
From a pattern recognition point of view, the resulting
classifier would be analogous to a decision list, a multi-classifier
system, or a decision tree (we will return to this point later).

This division into different types of control is intended to help
with discussions about timing behavior; we are not postulating that
there are different kinds of neural structures implementing
different kinds of control mechanisms.
Instead, the mechanisms controlling reconfigurability
are proposed to be somewhat analogous to motor control in the
general case (Types~V and~VI), but can perform simple and repetitive
types of control (Types~I through~IV) when faced with simpler tasks--often
the kind of constrained tasks encountered in psychophysical experiments.

The rest of the paper is mainly concerned with control mechanisms I - IV
and assuming simple, discrete reconfigurations in which  $\lambda_r\in\{0,1\}$
and $\sum \lambda_r =1$.
This should not be taken to mean that there is any reason
why more complex forms of control could not exist.
But we will see that these simple and limited forms of recurrence are
sufficient to explain many complex phenomena.

\section{Neurophysiological Basis}

Merely gaining some computational power through postulating more complex
computations than linear threshold units would by itself be unremarkable.
What makes Equation~\ref{eq-multiplexed} interesting is that it has a
simple, biologically plausible implementation in terms of real neurons.

Recall that most neurons (including most cortical neurons) consist of
a (usually) large dendritic tree, having somewhere between $10^4$ and $10^6$
synapses, a trigger zone, and an axon that eventually branches out
and forms further synapses.
The dendritic tree is a large branching membrane that has a topological
structure corresponding to a tree structure in mathematics and
computer science.
Signals transmitted by synapses can be excitatory or inhibitory
and generally contribute additively to the voltage at the trigger zone
of a neuron.
Some synapses can be multiplicative in nature.
Such multiplicative synapses
are linked to the topological structure of the tree.
That is, they combine multiplicatively with the additive contributions
of the synapses within their own branch.

When we combine additive and multiplicative synapses,
potentially arbitrary boolean expressions can be implemented,
with the tree structure of the dendritic tree corresponding
to the nesting of the boolean expression\cite{KochDendritic}.
A criticism of using the computational capabilities of
dendritic trees in this way is that it is unclear how
such structures can be created during ontogeny, or
even how such connectivity might be encoded genetically.

For reconfigurable networks, we take advantage of these
computational capabilities of dendritic trees, but
without requiring the kind of specific connectivity
needed for the implementation of arbitrary boolean circuits.
Instead, we rely on the existence of a multiplicative
synapse type that preferentially forms on branches near
the root of the dendritic tree.

Putting these observations together, we then arrive
at the following process for the development of reconfigurable
neural networks.
First, neurons from the area controlling the reconfigurable
pathway send axons into the pathway.
Each axon makes random multiplicative connections to
branches of the dendritic trees of neurons in the
reconfigurable pathway.

For the simple model described by Equation~\ref{eq-multiplexed},
there should be a connection from each controlling
neuron to only a single branch of each neuron in the
target pathway, and multiple controlling neurons should not
connect to the same target branch (more complex forms
of control and reconfiguration are possible, however).
Such constraints do not require specificity in the connection
and can be implemented in terms of simple signaling molecules
or development dependent on neural activity during development.

Alternatively, axons from controlling neurons can first
establish synapses on the body of neurons or on a dendritic
stump  in the target pathways, and this synapse formation
then itself triggers the formation of a larger dendritic
subtree under the control of the controlling neuron.

Through either mechanism, each neuron in the target pathway
then ends up with a collection of dendritic subtrees, each of which is
gated by a signal from a controlling neuron via a multiplicative
synapse.  Furthermore, the establishment of this kind of network
was based on random connections from controlling to target neurons
and requires only simple, local signaling during development.  
This distinguishes
it from many prior proposals for taking advantage of multiplicative
synapses

A second observation is that, except for the formation of the
controlling synapses, the neurons in the reconfigurable pathway
are free to form other synapses by whatever rules, statistics,
and signals that are required for a particular function.
Activation of a set of controlling synapses will simply select
a subset of these connections for a computation.
If these activation patterns for the controlling synapses are
disjoint, then the reconfigurable pathway will behave as if it
could be switched between multiple, unrelated feed forward
networks, as described by Equation~\ref{eq-multiplexed}
(actually, there is nothing in this description that prohibits
the formation of recurrent connections, so recurrent networks
can also be reconfigured using this mechanism).

In addition, signals flowing backwards from the body of the
neuron towards the synapses are also
blocked by the kind of multiplicative synapses used in this
construction.  This means that the reconfiguration implied
by the control signals does not just establish a multiplexing
for feed-forward computations, but also for learning.
That is, the reconfigurable pathway behaves in each configuration
as if it were a separate neural network, both for the purposes
of computation and for the purposes of learning.

\section{Evolutionary Rationale}

In the previous section, we saw that reconfigurable pathways
are easy to implement in terms of neural hardware:
the required computations and synapse types are common,
and creating the network only requires simple intercellular
signaling and random connections, both commonly observed
in the nervous system.
But although we have seen (and will explore in more detail below)
that reconfigurable pathways are useful for high-level computations,
the question is how such a complex mechanism might have arisen.

Let us consider a plausible evolutionary history of such
a pathway.
Neurons are energetically costly for organisms to maintain.
Reducing the number of neurons that an organism needs to
maintain is therefore evolutionarily potentially
highly advantageous.
The reconfiguration mechanism describe above permits this
by providing a simple mechanism for multiplexing a neural
pathway.
Such a circuit allows the same number of neurons to carry
out a much larger number of neural computations, where
the additional cost of adding new functions is limited
to the cost of adding the extra membrane needed for
the additional dendritic subtree, instead of developing
entirely new neurons.
The price paid for this is that multiplexed neural
computations need to be carried out sequentially instead
of in parallel.

Initially, control mechanisms may have been limited to
Type~II and~III control described above, which require
only a connection from a sensory input or a simple
oscillatory neural circuit to a pathway.
In fact, Type~II control, in which a neural pathway
is switched depending on sensory input, in its
simplest form is a simple kind of associative learning,
in which long term potentiation and/or long term depression
is conditioned on some other sensory input or context.

Once the necessary genetic circuitry for establishing
a control/reconfiguration relationship between
two neural areas was established,
the next step may have been increasing the complexity
of the control circuit, for example by reusing
neural circuitry for motor control and motor planning.
Once complex control of reconfigurable pathways was
available and provided greatly increased learning
and adaptation capabilities, it may then have become replicated
and formed the basis for cortical circuits in higher animals.

Although this evolutionary view is, of course, highly
speculative, it predicts that we should be able to find
early forms of reconfigurability already in simple
neural circuits, where they are potentially part of circuits
for long term potentiation, long term depression, and
classical conditioning.
It also predicts that some of these circuits would be
the predecessors of more complex reconfigurable pathways
in higher animals, suggesting that ontogenic signals
and neurotransmitters may be related between such simple
circuits and corresponding circuits in higher animals.


\section{Interpretation of Existing Experimental Results}

The previous sections introduced a biologically plausible
mechanism for extending the computational capabilities of
feed forward networks (we will return to this point in a later
section).
However, by itself, that is not particularly interesting,
since there are many such possible extensions.
What actually motivated the development of the above
architecture was a number of observations in neurophysiology
and psychophysics that were different to explain in
terms of proposed existing mechanisms.
This section describes two observations that motivated
the proposal of reconfigurable pathways: the observation
of stochastic and distributed representations in neurophysiology
and priming behavior in psychophysics.
The intent of giving these examples is not to provide a complete
theory or explanation of these phenomena in terms of reconfigurability
(that will require a much more detailed formulation of models
and examination of experimental results than possible here), but
merely to illustrate the kinds of explanations that reconfigurable
models can potentially yield for complex phenomena in
neurophysiology and psychophysics.

\subsection{Neurophysiology}

Neurophysiology has given insights into neural codes
in some cases, such as early visual representations\cite{VisualCortex}
and motor behavior like bird songs\cite{FeeBirds}.
In the case of early visual representations, neural codes
roughly correspond to common image transformations based on
simple statistical principles.
In the case of motor behavior, neural codes represent
spatial and temporal activation patterns for muscles.
In both cases, representations are largely determined
by the physical statistical nature of the signals.

However, such easily interpretable representations have
not been identified for higher level concepts.
For example, even though object categories and instances
appear to have psychological reality, no individual neurons
that clearly correspond to these concepts, so called ``grandmother
cells'', have been identified.

Another common observation in neurophysiological experiments
is that the response of neurons to stimuli tends to be
stochastic; that is, a given stimulus does not reliably
produce an action potential.

Since a simple correspondence between psychological concepts
and neural activity has not been identified, {\em stochastic,
distributed representations} have been postulated as representations
of psychological concepts such as object class and identity.
In such representations, no neuron responds specifically to
a concept or category, but instead concepts are encoded in
the joint activity of many neurons.
For object recognition, this means that instead of observing
cells that respond to specific objects, we expect to observe
cells that respond stochastically to multiple objects.
Such cells are referred to as {\em totem pole cells}\cite{ThorpeGrandmother}.

Intuitively, such representations are attractive because if
a system is based on stochastic and distributed representations,
we would expect it to be robust to the failure of individual neurons.
However, stochastic, distributed representations of the form
postulated for neural representations otherwise have not been
found much use for computational purposes.
Furthermore, little other support for their existence has
been found in the literature.

An additional problem with stochastic, distributed representations
is that they would seem to require a significant amount of temporal
integration in order to decode, for example into a motor response.
However, this is difficult to reconcile with experimental results
on fast feed-forward object recognition.

Reconfigurability provides a simpler explanation of these observations.
\begin{itemize}
\item The response of individual neurons appears stochastic under some
        experimental conditions because
        their responses are different in different configurations.
        If the configuration input to the pathway were known,
        the responses of neurons would be much more predictable
        relative to that input.
\item The response of neurons to multiple unrelated stimuli (``totem
        pole cells'') is explained not as distributed coding, but
        instead as different functions of the neurons in different
        configurations.  Within each configurations, the response
        of individual neurons is specific.
\end{itemize}

Let us note that context-dependent changes in the preferred stimuli
for neurons has been observed in {\em place cells}\cite{MoserPlace}.
Within a single ``room'' or context, place cells respond quite
specifically to spatial locations.
But when the room or context is changed, the same place cell
may respond to a different, unrelated location.
That is, we observe a neuron responding to a high level pattern
(in this case, location within a room), yet that response
varies and is context dependent.
Reconfigurable pathways in the above sense are a potential
mechanisms for place cells.

Finally, observe that some degree of reconfigurability is also
observed physiologically through fMRI (functional magnetic
resonance imaging).
That is, depending on the task being solved, different
brain areas are differentially active.
It has been demonstrated that changes in neural activity cause the
physiological changes observed using fMRI
(as opposed to the physiological changes causing changes in
neural activity).
That is, in some way, either neural output from a source area
or neural input to a target area is modulated in such a way
that the target area becomes more or less
involved in solving a mental task.

The question is now whether we can experimentally
distinguish reconfigurable pathways from alternative explanations
of the observed neural activity, such as explanations involving
Bayesian message passing or stochastic distributed coding.
A sufficient experimental result allowing us to distinguish
between the reconfigurability hypothesis and such other
explanations would be the observation of correlations between
signals going into axo-dentritic ``gating'' synapses near the cell body in
an area and the response of the cells.

The reconfigurability hypothesis predicts that if there
are multiple such incoming gating signals and we consider
the response of a cell only while a specific gating input is
active, the neuron will then behave in a more deterministic and more
specific way; that is, in the context of incoming gating signals,
it will stop behaving like a stochastic totem pole cell, and more
like a deterministic grandmother cell.

A second neurophysiological prediction is that synapses that
are suitable for performing the gating function for implementing
a reconfigurable pathway should respond to stimuli based on
context and priming (see below).  That is, certain contexts
should activate such synapses more frequently, while others
should depress them.

\subsection{Priming}

In the previous section, we looked at possible explanations
for observations in neurophysiology based on a reconfigurable
pathway model.
Let us now look at an example from psychophysics.

Priming is a process observed in many experiments in which
presentation of one stimulus, or context, affects the perception
of subsequent stimuli.
Most commonly, priming with a stimulus speeds up recognition
of subsequent, related stimuli.
Priming also reduces neural activity for subsequent
processing, as measured by neurophysiology,
fMRI, and EEG. 

Such results are generally interpreted in the framework of
a dynamic systems view of neural networks, typified
by Hopfield networks.
Within a dynamic systems framework, neural computations
need to converge to a final state and the time
for convergence depends on the state of the network
and the starting point for the computation.
However, despite this general view of priming,
no specific neural network models of priming are
widely accepted, and priming effects are difficult
to explain with commonly used feedforward
neural networks.

Reconfigurable pathways provide a alternative,
simple explanation for priming effects.
For computation with reconfigurable pathways,
the time until a result is obtained depends on
the number of configurations that need to be explored
and potentially the order in which they are
explored;
when a perceptual task can be solved by utilizing fewer
configurations, or a configuration that solves the
task can be identified earlier (depending on which
type of control we assume), the task can be completed
more quickly.
An explanation for priming effects in a
reconfigurable model is that they change
the order and/or choice of configurations used
in the performance of a task.

As a concrete example, consider a visual object recognition
task in which objects are presented sequentially on a display.
In a reconfigurable model of visual object recognition,
recognition would be carried out by many different
configurations of the visual pathways, where each configuration
recognizes a category of objects related by
invariance properties, degree of shape variation,
relative importance of color vs. shape, etc.
Priming with an image or concept can then be understood as
a reordering in which the different
configurations are considered, with configurations related
to the priming stimulus being considered earlier than they
otherwise would be.
Assuming Type~IV control (the simplest form of the
control mechanisms above that is capable of showing
variable timing), this would result in the experimentally
observed influence of priming on recognition speed.

This interpretation is also consistent with
neurophysiological results showing reduced and ``more
focused'' activity in the presence of priming.
Generally, neural computations at any one time only
activate a small fraction of all neurons in a pathway.
If two configurations represent very different kinds of computations,
the probability that a neuron is activated in one and the other
configuration would be independent of each other.
Therefore, the probability that any specific neuron is
activated throughout the performance of a task increases
with the number of configurations used in the task,
consistent with what is measured in priming studies.

It is an interesting question whether we can test and distinguish
reconfigurable models from other models purely using psychophysical
experiments.
Reconfigurability actually makes fairly specific predictions:
the amount of time required to complete a perceptual task grows
with the number of configurations that need to be considered
in its solution, and the accuracy depends on the best configuration
that was identified.
Experimentally, there are a number of mechanisms we can potentially
use to manipulate and measure the number of configurations that
are used in the solution to a task.
For example, if we give the subject only a fixed amount of time
to complete a task, the subject would ideally choose a statistically
optimal subset of configurations for solving that task.
As a special case, for only very short amounts of time for solving a task,
only a single configuration may be available.
Based on such manipulations, it should be possible to determine
whether experimentally observed timing and error behavior is
consistent with the predictions of reconfigurability.

It needs to be born in mind, however, that existing explanations
of timing and error behavior in psychophysical experiments tend to
be so non-specific that they are difficult to disprove by any experiment.
For example, models that are based on message passing or dynamical
systems views of neural systems generally allow for highly variable
response times for perceptual tasks in psychophysical experiments, but
they do not make many specific, falsifiable predictions.

In other areas of psychophysics, such as visual search, models based
on concepts such as ``parallel search'' and ``serial search'' are
commonly used in the interpretation of experiments.  Such models
are, however, merely descriptive of the timing behavior, not
mechanistic; that is, rather than being an alternative hypothesis
to reconfigurability, reconfigurability may provide a neural and
mechanistic basis for these timing behaviors.

On the whole, we can be optimistic that psychophysical experiments
can provide evidence for reconfigurable models.
At the same time, it is probably time for models based on dynamical
systems, message passing, and stochastic and distributed representations,
to be made concrete to the point where they can also make specific
timing and error predictions for different experimental conditions.

\section{Statistical Learning}

Learning and classification in feed-forward networks
is a well-researched and well-understood problem.
Computationally, feed-forward networks can approximate
functions, implement classifiers, implement boolean
circuits, and perform feature extraction.

Particularly important applications of feed-forward
neural networks are their use as classifiers.
Commonly used architectures and training techniques
for classifiers
are multilayer perceptrons and convolutional networks
trained with backpropagation,
radial basis functions trained using least square methods,
and support vector machines.
Nearest neighbor classifiers can also be implemented
easily with feed forward networks.

However, many classifiers that are commonly used
in pattern recognition, are less naturally implemented
in terms of neural networks.
Below, we discuss this for decision lists,
decision trees, multi-classifier systems, and
style models, showing different capabilities of
reconfigurable networks.

Such classifiers have in common that they are made up
of a potentially large, training-data dependent number
of component classifiers that are invoked in sequence
and whose outputs are integrated into a final
classification decision.
Let us call an abstract model of this class of classifiers
a {\em sequential compound classifiers}.
Such classifiers defined by a finite state
machine determining which classifier to execute next, how
to integrate the output of this classifier with previous
classifiers and when to terminate the computation.
The transition matrix is a function of the output of the
current classifier.
We will study these kinds of systems and their learning
algorithms more formally in a separate paper.

Although, in principle, many of these classifiers
can easily be parallelized, by implementing their component
classifiers in parallel and then combining their
outputs within another feed-forward, this is not a natural
way of implementing them, since the number (and
sometimes kind) of classifiers is data-dependent
and grows with more training data.
Furthermore, many of the component classifiers are
invoked only rarely, making dedicating neural hardware
to them wasteful.
Parallel implementations would require mechanisms
for allocating new groups of neurons as more
input data becomes available and connecting these
new groups back to existing groups in the right way.
Furthermore, the time required for classification
in such classifiers depends on the input data.

Reconfigurable pathways permit a more direct implementation
of these kinds of classifiers:
different component classifiers are implemented by
different configurations, and the order
in which they are invoked and the way their
outputs are combined is handled by the control circuits.
Let us look at these kinds of classifiers in more detail.

For {\em decision lists}, a sequence of classifiers needs
to be executed.  Each such classifier outputs either the
final classification, or a special class indicating that
it could not provide the answer.
The control circuit circles through a sequence of states
until the first classifier returns a final classification.
Such a control circuit can be implemented as a simple
feed forward network in terms of linear threshold units.

Let us look at this construction in more detail in the
case of decision lists to illustrate how this works
mathematically.
For a decision list, we have a sequence of classifiers
$f_i$ giving some output $y = f_i(x)$, where the $f_i$
are implemented as configurations of the
pathway and feedforward networks.
As component classifiers of a decision list classifier,
each classifier either outputs a classification, or an
indication that it couldn't classify the input sample;
we assume that this indicator is a binary output $y_0$, with
$y_1\ldots y_n$ representing classification output.
We augment the output of each configuration with a set of
neurons representing a unary encoding of the configuration
number $l_j = \delta_{ij}$, with
$\delta_{ij} = \left\lfloor i=j \right\rfloor$.
The control network is itself a feed-forward classifier that,
as input, takes $y_0$ and the $l_j$ and produces as output
a unary representation $\lambda_k$ of the next configuration
to be used.
Without the stopping criterion, we would simply have
a feed forward control network that cycles through the
configurations (here, $(x)_R$ is a shorthand for $x \mod R$.
$$\lambda_k = l_j \cdot \delta_{k,(j+1)_R}$$
where $R$ is the total number of configurations
With the stopping criterion, we arrive at:
$$\lambda_k = (1-y_0)\cdot l_j \cdot \delta_{k,(j+1)_r} + y_0 \delta_{k,0}$$
To guard against noise, we may want to apply a threshold unit
instead of using the linear computation.

In this construction, we have considered the state labels $l_j$
an augmentation to the feed forward networks in the reconfigurable
pathways.  This lets us consider the entire pathway and its control
network to be a single, simple feed forward network where the only
information that flows recurrently is the choice of configuration $r$.
Anatomically, the state labels $l_j$ might, of course, also
be located in the control network, giving the control network
the appearance of a more complex recurrent network.

For {\em decision trees}, there are two kinds of classifiers,
those that are in the interior of the tree, and those that
are at the leaves.  The former return as output the next
classifier to be tried, the latter return a final
answer to the classification problem.
The control circuit is similar as for decision lists,
but it needs to keep track of the relationship between
classifiers, classifier outputs, and states.

{\em Multi-classifier systems} compute the
outputs from multiple individual classifiers
and then combine the outputs of the discriminant functions
or posteriors of the classifiers, often by averaging.
In the simplest case, the control circuit cycles through a
static sequence of component classifiers, although more
complex choices are possible.
The control circuit also needs to coordinate the averaging
of the classifier outputs; alternatively, the outputs
may simply be averaged using time constants,
so that the final output of the pathway is always a
weighted temporal average of the outputs of the past
several configurations in the pathway.

{\em Style models} are models in pattern recognition in which
a group $g$ of related classification problems $\{x_{g,1},\ldots,x_{g,r}\}$
can be solved better by taking advantage of their relationship.
One of the most general models for styles is the hierarchical
Bayesian model, in which we assume that the class conditional
densities $p(x|c,\theta)$ are dependent on some parameter $\theta$,
and for a group of related samples, $\theta$ was sampled from
a prior parameter density $p(\theta)$.
When the prior parameter density is a linear combination of
delta functions,
$$p(\theta) = \sum_i \Lambda_i \delta(\theta,\theta_i)$$
then the hierarchical Bayesian model becomes a finite mixture model:
$$p(x|c,\theta) = p(x|c,\lambda) = \sum _i \lambda_i p_i(x|c)$$
Often, the $\Lambda_i$ are assumed to be in $\{0,1\}$ (i.e.,
the world is in one of a number of discrete and distinct states
for each group of input samples), and hence a maximum likelihood
estimate for the $\lambda_i$ is itself in $\{0,1\}$.
If we take a Bayesian approach to classification, the $\lambda_i$
take on continuous values.
For optimal classification under a zero-one loss function, the
classifier should classify using the posterior probabilities
(or a strictly monotonic function thereof) as the discriminant
function.  The posterior probabilities are easily seen to be
expressible in the form:
$$p(c|x) = \sum_i \lambda_i p_i(c|x)$$
This is similar to the multi-classifier case, but the final output
is a weighted average, not just an average, of the different
component classifiers.

The weights themselves are determined by how well each group $g$
of samples is explained by each model.
The optimal decision is also a classification problem, namely
determining whether $m_g$, the model for group $g$, is given
by the set of class conditional densities $p_i(x|c)$.
In the absence of class labels or other class-related information,
we can only determining which model fits the data in the group
best by marginalizing and considering $p_i(x)$.
We then obtain using Bayes formula and assuming independent samples:
$$p(m_g=i|G) = p(m_g=i|x_{g,1},\ldots,x_{g,r}) \propto \prod p_i(x_{g,i}) P(m_g=i)$$
The weights $\lambda_i$ should be chosen either proportional to
$p(m_g=i|G)$ or (for maximum likelihood) as $\arg\max_i p(m_g=1|G)$.
(If extra information is available prior to the presentation of the
training samples, that gives us information about $P(m_g)$, then
we have a situation related to priming in psychophysical experiments.)

Since neural systems usually face a sequential decision problem, in
which the first elements of a group $G$ need to be classified before the
rest of the batch is available, and in which transitions between different
groups may not be marked explicitly, a fairly simple strategy for
estimating the $p(m_g=i|G)$ is to approximate it with a sliding
window, i.e. for a given sample $x_n$:
$$\hat{p}(m_g=i|G) \approx \prod_{j=n-k}^n p_i(x_j)$$
If we take logarithms on both sides, we obtain:
$$\log\hat{p}(m_g=i|G) \approx \log\sum_{j=n-k}^n \log p_i(x_j)$$

This then gives a simple prescription for style models based
on reconfigurable pathways:
\begin{itemize}
\item Each configuration of the pathway is a classifier that estimates
	$p_i(c|x)$, as well as a density estimator that estimates
	$\log p_i(x)$ (this can be interpreted as a confidence
	the classifier has in classifying the input sample)
\item The pathway is cycled through all relevant configurations $i$
    (configurations for which $\log p_i(x)$ is low during one
    iteration may simply be skipped on future iterations)
\item The outputs are integrated by temporally integrating the $\log p_i(x)$
    and averaging the $p_i(c|x)$ using as weights the exponential
    of the integrated weights.
\end{itemize}

In this way, we see that reconfigurable pathways can implement
a common form of hierarchical Bayesian classifiers (namely, those
represented as mixture models), even for large numbers of mixture
components.
Let us note that if we drop the constraint that neural pathway
configurations are totally distinct, and assume instead that
mixed configurations are possible, we can imagine more complex
hierarchical Bayesian models, in which the control of the
pathway attempts to maximize $p(x)$ on average by ``mixing''
different configurations together.

Overall, the goal of this section has been to show that reconfigurable
pathways permit the plausible implementation of a rich set of
widely used classifiers and pattern recognition methods that, due
to their sequential nature, used to be considered not plausible
as instances of neural computation.

\section{Discussion}

The paper has described a biologically plausible mechanism that
allows a collection of neurons or an entire
neural pathway comprising multiple areas and connections between
them, to be reused for multiple different, potentially unrelated
computations.
We have seen evolutionary rationales, and some relationships to findings
in neurophysiology, psychophysics and machine learning.
Of course, the discussions in this paper are not sufficient to
prove the actual use of reconfigurable pathways in the brain.

Nevertheless, the reconfigurability hypothesis here has
been stated in greater detail and with more experimental predictions
than other common theories of dynamic brain function and distributed
representations, such as Bayesian message passing, dynamic systems theories of 
brain function, and stochastic distributed representations.
In particular, the reconfigurability hypothesis makes specific
predictions about the kinds of neural signals and circuits we
expect to find neurophysiologically, as well as predictions about
the relationship between task structure and timing across potentially
a wide range of psychophysical experiments.

The analysis of psychophysical and neurophysiological results
above can, of course, only be considered tentative.
The next steps are to look in much more detail at the degree
to which existing results can be explained in terms of reconfigurability,
as well as to devise specific experimental tests of the
reconfigurability hypothesis.

From a theoretical and machine learning point, the
reconfigurability mechanism is significant, not so much in that
it contributes new machine learning methods, but rather in that
it provides a plausible neural basis for a large range of
existing machine learning methods (e.g., decision trees, decision lists,
hierarchical Bayesian methods), for which previously specific
neural implementations either didn't exist or even seemed
implausible.

No matter whether reconfigurability will ultimately turn out to
true, it does provide a challenge to common interpretations of
neurophysiological and psychophysical results.
Right now, these experiments are generally interpreted with the
implicit assumption that neurons and brain areas perform specific,
relatively consistent functions, and that properties of, and computations
performed by, say, the visual system remain stable across experiments.
The mere possibility that different experiments may actually
be testing completely different neural computations is a significant
factor that needs to be taken into account in future work.

\bibliographystyle{plain}
\bibliography{reconf}

\end{document}